\documentclass{bmvc2k}

\title{UBR$^2$S: Uncertainty-Based Resampling and Reweighting Strategy for Unsupervised Domain Adaptation}

\addauthor{Tobias Ringwald}{tobias.ringwald@kit.edu}{1}
\addauthor{Rainer Stiefelhagen}{rainer.stiefelhagen@kit.edu}{1}

\addinstitution{
 Institute for Anthropomatics and Robotics (CV:HCI Lab)\\
 Karlsruhe Institute of Technology\\
 Karlsruhe, Germany
}

\runninghead{Ringwald \etal}{UBR$^2$S for Unsupervised Domain Adaptation}

\def\eg{\emph{e.g}\bmvaOneDot}

\def\etal{\emph{et al}\bmvaOneDot}

\usepackage{epsfig}
\usepackage{graphicx}
\usepackage{amsmath}
\usepackage{amssymb}

\usepackage[page,toc]{appendix}

\usepackage{bbding}
\usepackage{pifont}
\usepackage{wasysym}
\usepackage{amsmath}
\usepackage{mathtools}
\usepackage{amssymb}
\usepackage{array}
\usepackage{arydshln}
\usepackage{multirow}
\usepackage[ruled,vlined,noend]{algorithm2e}
\usepackage{booktabs}
\usepackage{adjustbox}
\usepackage{color, colortbl}

\newcommand\Tstrut{\rule{0pt}{2.3ex}}
\def\wrt{w.r.t\bmvaOneDot} 

\newcommand{\z}[2]{#1{\small $\pm$#2}}

\begin{document}

\maketitle

\begin{abstract}
   Unsupervised domain adaptation (UDA) deals with the adaptation process of a model to an unlabeled target domain while annotated data is only available for a given source domain. This poses a challenging task, as the domain shift between source and target instances deteriorates a model's performance when not addressed. In this paper, we propose UBR$^2$S -- the Uncertainty-Based Resampling and Reweighting Strategy -- to tackle this problem. UBR$^2$S employs a Monte Carlo dropout-based uncertainty estimate to obtain per-class probability distributions, which are then used for dynamic resampling of pseudo-labels and reweighting based on their sample likelihood and the accompanying decision error. Our proposed method achieves state-of-the-art results on multiple UDA datasets with single and multi-source adaptation tasks and can be applied to any off-the-shelf network architecture. Code for our method is available at \url{https://gitlab.com/tringwald/UBR2S}.
\end{abstract}

\section{Introduction}

Modern convolutional neural networks (CNNs) require the optimization of millions of parameters by learning from a vast amount of training examples \cite{imagenet}. While this training data might be readily available for a given source domain, annotated data for the actual domain of interest -- the target domain -- might be nonexistent or very hard to obtain. For example, synthetic images could be generated en masse and used for training, while classifying unannotated real world data (\eg medical images) is the actual objective. Unfortunately, the domain shift between the source and target data results in a severely degraded performance when utilizing current classification approaches.

\begin{figure}[t!]
\vspace{0.25cm}
\begin{tabular}{c}
\bmvaHangBox{%
\includegraphics[width=0.9\textwidth]{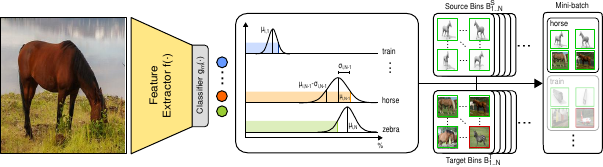}%
}
\end{tabular}
\vspace{0.25cm}
\caption{Overview of our proposed resampling strategy. Before each adaptation cycle, the current state of the network is frozen and used to extract class-wise uncertainty distributions quantified by Monte Carlo dropout. Based on these uncertainty distributions, class scores are resampled and converted into a pseudo-label. The instances are then grouped together with similar instances into bins, which are later used for sampling mixed mini-batches containing both source and target examples.}
\label{fig:overview_fig}
\end{figure}

Unsupervised domain adaptation (UDA) seeks to address the domain shift problem under the assumption that no annotated data for the target domain is available. Prior work in this area approached the problem from several different angles: Image and pixel-level methods were proposed for learning a direct image-to-image mapping between the different domains, thereby enabling the transfer of target data into the source domain (or vice versa), where straightforward training is then feasible~\cite{hoffman2017cycada,atapour2018real}. At the feature-level, UDA methods often rely on minimizing common divergence measures between the source and target distributions, such as the maximum mean discrepancy (MMD) \cite{can}, Kullback-Leibler divergence~\cite{meng2018adversarial} or by enforcing features from both domains to be indistinguishable with the help of adversarial training~\cite{hoffman2017cycada,simnet}. However, these approaches often require complicated training setups and additional stages such as domain discriminators \cite{simnet}, gradient reversal layers~\cite{ganin2014unsupervised} or other domain-specific building blocks~\cite{dsbn}. This adds both millions of parameters to be optimized and also further hyperparameters that have to be tuned. In this paper, we instead rely on a model's inherent prediction uncertainty for the unsupervised domain adaptation task, which we quantify by Monte Carlo dropout~\cite{mcdropout}. Our proposed method leverages the extracted uncertainty for dynamic resampling, assignment and reweighting of pseudo-labels and can be directly applied to any off-the-shelf neural network architecture without any modification. Furthermore, we propose domain specific smoothing (DSS) -- a label smoothing based improvement to the training pipeline for UDA setups.

To summarize, our contributions are as follows: (i) We propose UBR$^2$S -- the uncertainty-based resampling and reweighting strategy utilizing a model's prediction uncertainty under Monte Carlo dropout. (ii) We introduce DSS -- the domain specific smoothing operation for UDA tasks. (iii) We evaluate our method on multiple common UDA benchmarks and achieve state-of-the-art results on both single and multi-source adaptation tasks. (iv) We show that UBR$^2$S works with a plethora of network architectures and outperforms recent methods while using only a fraction of their parameters.

\section{Related Work}

Unsupervised domain adaptation has been the subject of many prior works in recent years and was addressed in multiple different ways. The authors of~\cite{atapour2018real} leverage GAN-based style transfer in order to translate synthetic data into the real domain and achieve domain adaptation for a monocular depth estimation task. Similarly, Deng ~\etal~\cite{deng2018image} apply unsupervised image-to-image translation for their person re-identification task and consider the self-similarity and domain-dissimilarity of source, target and translated images.
Instead of aligning domains at the image-level, prior methods have also considered alignment at the feature-level: One of the first works in this direction was the gradient reversal layer (\textit{RevGrad}) proposed by Ganin~\etal~\cite{ganin2014unsupervised}. They achieve domain-invariant feature representations by forcing the feature distributions from the source and target domain to be as indistinguishable as possible with a domain classifier. Related to this, Pinheiro~\cite{simnet} proposes a similarity-based classifier that combines categorical prototypes with domain-adversarial learning. Park~\etal~\cite{park2018adversarial} show that training with their proposed adversarial dropout can also help to improve generalization.
In place of adversarial training, distribution divergence measures have also been applied successfully in order to align the source and target domain at the feature-level: Long~\etal~\cite{long2013transfer} use the maximum mean discrepancy (MMD) measure for alignment, which recently was extended by Kang~\etal~\cite{can} for their proposed CDD loss and also used in the regularizer proposed by Gholami~\etal~\cite{gholami2017punda}. In a similar way, Meng~\etal~\cite{meng2018adversarial} utilize the Kullback-Leibler divergence as another distribution difference measure.
Recently, Hoffman~\etal~\cite{hoffman2017cycada} have combined both feature- and image-level approaches in their proposed CyCADA framework, which adapts feature representations by enforcing local and global structural consistency and also employs cycle-consistent pixel transformations. 
In terms of pseudo-labeling target instances, Saito~\etal~\cite{saito2017asymmetric} propose an asymmetric training method that consists of three separate networks where two networks act as pseudo-labelers and the third network is trained on said labels to obtain discriminative representations for the target domain samples. Related to this, Zhang~\etal~\cite{zhang2018collaborative} propose an iterative pseudo-labeling and sample selection approach based on an image and domain classifier. This idea is also picked up by Chen~\etal~\cite{chen2019progressive}, who employ their progressive feature alignment network and an easy-to-hard transfer strategy for iterative training.
Another direction was pursued by Chang~\etal~\cite{dsbn}, who propose the use of domain-specific batch-normalization layers in order to deal with the distribution shift between domains. This concept was also employed in the contrastive adaptation network (CAN) proposed by~\cite{can} and proved to capture the domain specific image distributions.

Most related to our work in this paper, Long~\etal~\cite{long2018conditional} have explored the use of uncertainty for their proposed CDAN architecture and achieve domain adaptation by controlling the classifier uncertainty to guarantee transferability between domains. Han~\etal~\cite{han2019unsupervised} quantify model uncertainty under a general Rényi entropy regularization framework and utilize it for calibration of the prediction uncertainties between the source and target domain. Gholami~\etal~\cite{gholami2017punda} consider the Shanon entropy of probability vectors in order to minimize a classifier's uncertainty on unlabeled target domain instances. Similar to the approaches above, Manders~\etal~\cite{manders2018adversarial} propose an adversarial training setup forcing prediction uncertainties to be indistinguishable between domains.
In this work, we explore the usage of prediction uncertainties quantified under the Monte Carlo dropout~\cite{mcdropout} approximation of Bayesian inference. Unlike prior work, our proposed UBR$^2$S method leverages a model's prediction uncertainty for dynamic resampling, assignment and reweighting of pseudo-labels. Our method does not require any image- or feature-level adjustments and can thus be applied to any off-the-shelf neural network. Nevertheless, it still achieves state-of-the-art results and is also competitive when using smaller feature extractors with a fraction of the usual parameters.

\section{Methodology}

Unsupervised domain adaptation (UDA) seeks to address the domain shift between a source and target domain in order to maximize a model's generalization performance on the target domain while only given annotated data for the source domain. Formally, the annotated source dataset $\mathcal{D}_\mathcal{S}$ consists of input-label pairs $\{x_i^s, y_i^s\} \in \mathcal{D}_\mathcal{S}$ while the target dataset $\mathcal{D}_\mathcal{T}$ only contains unlabeled inputs $\{x_i^t\}$. Labels $y_i^s$ are elements of class set $\mathcal{C}=\{1, 2, \cdots, N\}$ with $N$ classes. Given this definition, the objective of UDA tasks is to produce accuracte predictions $y_i^t$ for every input $x_i^t$ of the target domain dataset.

The method discussed in this paper is presented in the context of deep neural networks that consist of a convolutional neural network (CNN) feature extractor $f(\cdot)$ followed by a classifier $g(\cdot)$ that projects $f$'s output into a probability distribution over the class set $\mathcal{C}$. 
A key part of our proposal is the concept of uncertainty quantified by Monte Carlo dropout (MCD)~\cite{mcdropout}, which we will now formally introduce. Let $\mathcal{M}$ be a set of size $\lvert \mathcal{M}\rvert$ containing binary masks $m_{1..\lvert \mathcal{M}\rvert}$ sampled from a Bernoulli distribution according to the Monte Carlo dropout rate, where $\lvert \mathcal{M}\rvert$ represents the number of MCD iterations. We then evaluate all dropout-masked classifiers $g_{m \in \mathcal{M}}(x_i^t)$ for a given target domain sample $x_i^t$ and quantify its class-wise uncertainties by the mean $\mu$ and standard deviation $\sigma$ for every class $c \in C$ as follows: 

\begin{equation}
    \mu_c(x_i^t) = \frac{1}{\lvert \mathcal{M}\rvert} \sum_{m \in \mathcal{M}} \left[ g_{m}\left(f\left(x_i^t\right)\right)\right]_c,\quad \sigma_c(x_i^t) = \sqrt{\frac{1}{\lvert \mathcal{M}\rvert-1} \sum_{m \in \mathcal{M}} \left[g_{m}\left(f\left(x_i^t\right)\right]_c - \mu_c(x_i^t) \right)^2}  \label{eq:mcd_mustd}
\end{equation}
\newline
For the sake of clarity, $\mu_c(x_i^t)$ will be shortened as $\mu_{i,c}$ in the following paragraphs ($\sigma_{i,c}$ likewise). With these prerequisites, we will now introduce our proposed uncertainty-based resampling and reweighting strategy (UBR$^2$S).

\subsection{Resampling Strategy}\label{section:resampling_strat}
For unsupervised domain adaptation tasks, annotations are only available for the source domain. These are oftentimes used for model initialization via supervised pretraining and allow for initial pseudo-label estimates on the unannotated target domain. Due to the domain shift between source and target instances, these initial estimates are inherently noisy. Despite their noisy nature, recent research in this area \cite{dsbn,zhang2018collaborative} often relies on the class predicted with maximum probability score in order to generate a pseudo-label, thereby neglecting the possibility of other classes. 
For UDA tasks, however, a model's maximum probability prediction for target domain data often does not correspond with the ground truth class after the \textit{source-only} pretraining stage. 
Our resampling strategy will thus consider predictions other than the maximum for assignment of a pseudo-label.

Given $f$ and $g$ after supervised pretraining on the source domain dataset, we start by extracting uncertainty measures $\mu_{i,c}$ and $\sigma_{i,c}$ for the c-th class of the i-th target domain sample (see Equation~\ref{eq:mcd_mustd}). As this is a continuous distribution, we first resample the i-th target instance's probability scores as $\Tilde{p}_{i,c} \sim \mathcal{N}\left(\mu_{i,c}, \sigma_{i,c}\right)$ in order to obtain discrete values. Subsequently, $\Tilde{p}_i$ is re-normalized and then used for the assignment of a pseudo-label $\psi(\frac{\Tilde{p}_i}{\sum_j \Tilde{p}_{i,j}}) = \Tilde{y}_i$ where $\psi: \mathbb{R}^{\lvert \mathcal{C} \rvert} \rightarrow \mathcal{C}$ is the weighted random sample function. This thus enables the usage of classes with non-maximum prediction scores for pseudo-labels based on the model's own predictive uncertainty.
The resampling step ends by assigning the i-th target sample to the bin $\mathcal{B}^\mathcal{T}_{\nu_i}$ based on $\nu_i = \mathrm{argmax}_{c \in \mathcal{C}}\, \mu_{i,c}$.  
Here, bins are groups of similar samples that are later used for construction of mini-batches (see Section~\ref{section:training_loop}).

The above resampling process of our proposed method is also visualized in Figure~\ref{fig:overview_fig}.

\subsection{Reweighting Strategy}\label{sec:reweighting_strat}
\begin{figure}[t!]
\renewcommand{\arraystretch}{2.0}
\begin{tabular}{cc}
\bmvaHangBox{\includegraphics[width=0.47\textwidth]{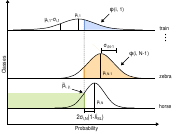}} & %
\bmvaHangBox{%
\begin{minipage}{.575\textwidth}
\centering
\begin{adjustbox}{width=.45\textwidth}
\hspace{-3cm}
\setlength{\interspacetitleruled}{0pt}%
\setlength{\algotitleheightrule}{0pt}%
\begin{algorithm}[H]
\SetAlgoLined
{\small \textbf{Input.} Source-trained $f$ and $g$ with weights $\theta$\;}
\For{$j=1$; $j \le T_\mathrm{cycles}$}{
    $\mu, \sigma \leftarrow \mathrm{extractUncertainty}(\mathcal{D}_\mathcal{T}, f, g, \theta)$\;
   \For{$\text{step}=0$; $\text{step} < T_\mathrm{steps}$}{
       \If{$\text{step}\, \mathrm{mod}\, 10 = 0$}{
            $\Tilde{p} \leftarrow \mathrm{resample}(\mu, \sigma)$\;
            $\Tilde{y} \leftarrow \Psi(\frac{\Tilde{p}_i}{\sum_n \Tilde{p}_{i,n}})$\;
            $\forall i: \nu_i \leftarrow \mathrm{argmax}_{c \in \mathcal{C}}\, \mu_{i,c}$\;
            $\forall i: \mathcal{B}^\mathcal{T}_{\nu_i} \leftarrow \mathcal{B}^\mathcal{T}_{\nu_i} \cup \{x_i^t\} \in \mathcal{D}_\mathcal{T}$\;
        }
        $c \leftarrow \mathrm{sampleClasses}(\mathcal{C}, \beta)$\;
        $b \leftarrow \mathrm{sampleBatch}(\mathcal{B}^\mathcal{S}, \mathcal{B}^\mathcal{T}, c)$\;
        $\lambda_\mathrm{SL}, \lambda_\mathrm{DE} \leftarrow \mathrm{calcError(\mu, \sigma, \Tilde{p}, \Tilde{y})}$\;
        $\theta \leftarrow \mathrm{train}(\theta, b, \lambda_\mathrm{SL}, \lambda_\mathrm{DE})$\;
    }
}
\end{algorithm}
\end{adjustbox}
\end{minipage}}\\
	(a) Graphical interpretation & \hspace{-2.5cm} (b) Training loop \\
\end{tabular}
\caption{(a) Graphical interpretation of our proposed reweighting step. Using resampling provides the opportunity to consider a non-maximum prediction at the cost of sampling a wrong pseudo-label. Our loss reweighting step assesses this risk based on the class-wise uncertainty distributions by calculating the decision error based on $\varphi(i, c)$ and sample likelihood $\lambda_\text{SL}^i$ for the i-th target instance in a N-class classification task. (b) Training loop of UBR$^2$S.}
    \label{fig:decision_error}
\end{figure}

Resampling a pseudo-label as described above allows for the consideration of non-maximum predictions. However, this is accompanied by the inherent risk of sampling the wrong class. After all, the maximum prediction is a reasonable estimate for some instances. We thus need to compensate for the potential decision error when choosing one class over another and also consider the likelihood of the current resampled value.
We calculate the sample likelihood (SL) as $\lambda_{\mathrm{SL}}^i = 1 - \left. \frac{\lvert \Tilde{p}_{i,\Tilde{y}_i} - \mu_{i,\Tilde{y}_i} \rvert}{2 \sigma_{i,\Tilde{y}_i}} \right\vert_0^1$ where $\cdot \big\vert_0^1$ clamps the value into range $[0, 1]$. Intuitively, this reflects the likelihood of the resampling step by measuring the deviation from the mean. However, it does not consider the risk involved with choosing the wrong class in the first place. For this reason, we determine this decision error based on the classes' uncertainty distributions by calculating the inverse of the cumulative probability $\Phi$ \wrt $\Tilde{p}_{i,\Tilde{y}_i}$ as per Equation~\ref{eq:basic}: 

\begin{equation}
    \varphi(i, c) = 1 - \Phi(\Tilde{p}_{i, \Tilde{y}_i}, \mu_{i,c}, \sigma_{i,c}), \,\, \text{with}\;\; %
    \Phi(x, \mu, \sigma) = \frac{1}{2} \left[ 1 + \mathrm{erf}\left(\frac{x-\mu}{\sigma \sqrt{2}}\right) \right] \label{eq:basic}
\end{equation}
\newline
Here, $\mathrm{erf}$ denotes the Gauss error function. 
The final decision error $\lambda_{\mathrm{DE}}$ is then given by Equation~\ref{eq:final_de} and calculated \wrt every class besides the current label estimation $\Tilde{y}_i$. The graphical interpretation of this procedure is visualized in Figure~\ref{fig:decision_error}a.

\begin{equation}
    \lambda_{\mathrm{DE}}^i = 1 - \mathrm{max}\left(\{\varphi \left(i, c\right) |\; \forall c \in \mathcal{C}\setminus \Tilde{y}_i\} \right) \label{eq:final_de}
\end{equation}
\newline
Finally, we normalize the product of $\lambda_{\mathrm{SL}}$ and $\lambda_{\mathrm{DE}}$ to a distribution with its center point at 1 and use it to dynamically reweigh the element-wise loss while training on target domain samples. Therefore, the loss contribution of a given sample depends on the certainty of the currently chosen pseudo-label. An in-depth description of this procedure will be given in Section~\ref{section:training_loop}.

\subsection{Domain Specific Smoothing} \label{section:dss}

This section will now introduce our proposed Domain Specific Smoothing (DSS) for UDA training. DSS is based on label smoothing, which was first proposed by Szegedy \etal \cite{labelsmoothing} and is commonly used to curb overfitting and overconfident predictions. In a normal $N$ class training setup, a label encoding function $v: \mathcal{C} \rightarrow \mathbb{R}^N$ would construct a discrete one-hot probability distribution so that one class is assigned 100\% with all other $N-1$ classes being at 0\%. With label smoothing, $v$ constructs a smoothed label vector by mapping a ground truth label (or estimated pseudo-label) $c \in \mathcal{C}$ into probability space according to Equation \ref{eq:labelsmoothing}. 

\begin{align}
  v(c)_i = \begin{cases}
    1 - \varepsilon, & c = i \\
    \frac{\varepsilon}{\lvert \mathcal{C} \rvert - 1}, & c \neq i
  \end{cases}
  \label{eq:labelsmoothing}
\end{align}

Here, $\varepsilon$ is the smoothing factor. When training with a cross entropy loss, probabilities are needed for the loss calculation. However, neural networks usually output unnormalized logits. Softmax normalization is thus applied to convert the logits into probabilities: $\vartheta(\ell)_i = \frac{e^{\ell_i}}{\sum_j e^{\ell_j}}$ where $\ell$ is the logit vector. Cross entropy loss is then given as $-\sum_i y_i\, \mathrm{log}\, \vartheta(\ell)_i$.
Let output logit vector $\ell = [l_0, l_1]$ and training target $y = [1.0, 0.0]$ be subject of a training step with 2 classes. The objective of training with cross entropy loss is the assignment of target class $c_0$ to 100\% and $c_1$ to 0\%. Because of softmax normalization, this can only be the case when logit $\ell_0\rightarrow\infty$ or $\ell_1\rightarrow-\infty$. Due to the lack of floating point accuracy, this happens before approaching infinity in reality: $\ell=[19, 0]$ already results in a $\vartheta(\ell)\approx [1.0, 0.0]$ assignment. However, when using label smoothed target $\overline{y}=[0.8, 0.2]$, logits $\ell=[2\ \mathrm{log}\ 2, 0.0]$ are already enough to match the target probabilities of $\overline{y}$ (with $\varepsilon=0.2$). As softmax is invariant to constant addition, $\ell=[q+2\ \mathrm{log}\ 2, q]$ leads to the same result for any choice of $q \in \mathbb{R}$. The absolute difference needed between $\ell_0$ and $\ell_1$ is therefore multiple times smaller for the smoothed target ($\frac{2\ \mathrm{log}\ 2}{19} \approx 0.07$). For training with target $y$, this has the consequence of rapidly growing weights prior to the output layer in order to boost the logit values and thereby minimizing the cross entropy loss. This, however, is a prime example of overfitting as stated by Krogh~\etal~\cite{krogh1992simple} and is also in conflict with Lawrence~\etal~\cite{lawrence2000overfitting} who found that smaller weights tend to generalize better. Thus, label smoothing can be seen as a regularization method that diminishes this adverse influence on training. 

Similar to prior UDA setups~\cite{zhang2018collaborative,can}, our method constructs mini-batches using instances from both the source and target domain. While the presence of ground truth source instances can diminish the effect of wrong target pseudo-labels, constant training on source data will make the model overly focus on this domain even though the adaptation to the target domain is the actual objective. This is in conflict with prior research, which indicates that good transferability and generalization requires a non-saturated source classifier~\cite{chen2019progressive}. We extend this idea to domain adaptation tasks and propose domain specific smoothing (DSS): With DSS, label smoothing is only applied to source instances, even when training with a mixed batch. This leads to a mixture of one-hot pseudo-labels for the target domain instances and smoothed ground truth labels for the source instances. As later shown in our experiments, this helps to improve the generalization performance after the pretraining on source domain data and also improves the domain adaptation capabilities of the final model.

\begin{figure}[t!]
\vspace{0.25cm}
\renewcommand{\arraystretch}{2.0}
\begin{tabular}{cccc}
\bmvaHangBox{
	\begin{minipage}[c][1\width]{0.21\textwidth}
	   \centering
        \includegraphics[width=0.5\textwidth]{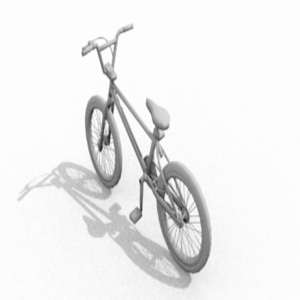}%
        \includegraphics[width=0.5\textwidth]{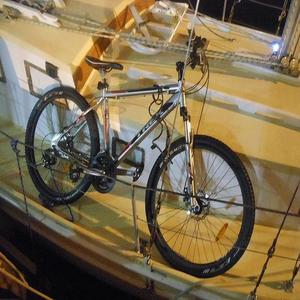}\\
        \includegraphics[width=0.5\textwidth]{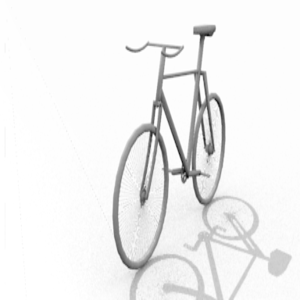}%
        \includegraphics[width=0.5\textwidth]{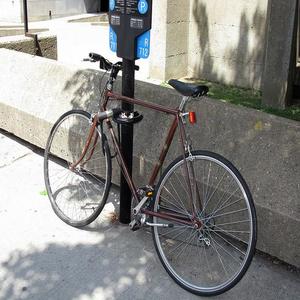}
	\end{minipage}} & \bmvaHangBox{
	\begin{minipage}[c][1\width]{0.21\textwidth}
	   \centering
        \includegraphics[width=0.5\textwidth]{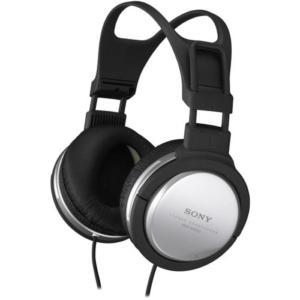}%
        \includegraphics[width=0.5\textwidth]{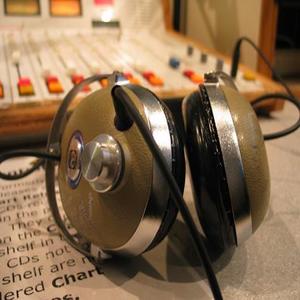}\\
        \includegraphics[width=0.5\textwidth]{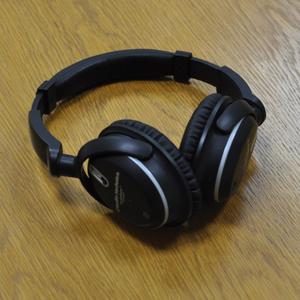}%
        \includegraphics[width=0.5\textwidth]{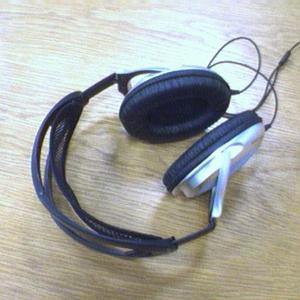}\\
	\end{minipage}} & \bmvaHangBox{
	\begin{minipage}[c][1\width]{0.21\textwidth}
	   \centering
        \includegraphics[width=0.5\textwidth]{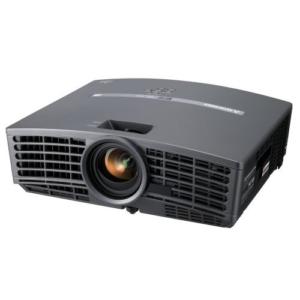}%
        \includegraphics[width=0.5\textwidth]{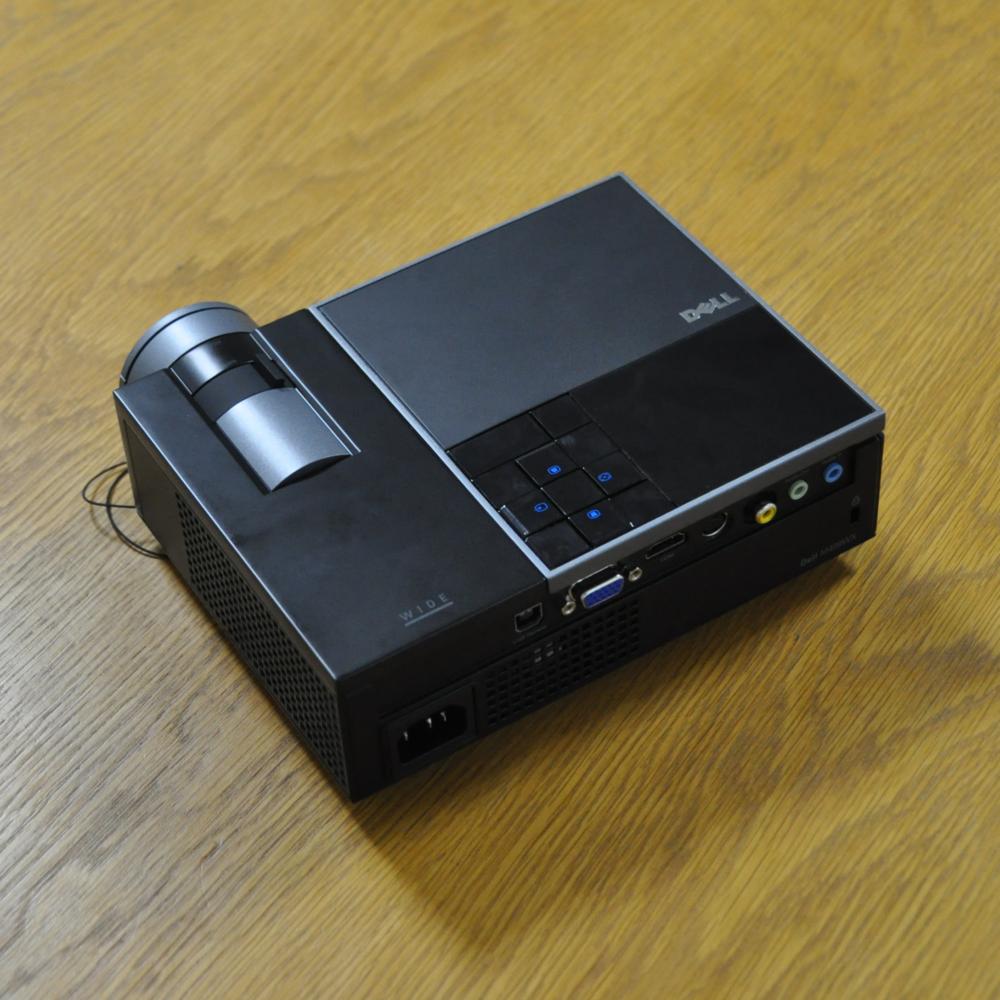}\\
        \includegraphics[width=0.5\textwidth]{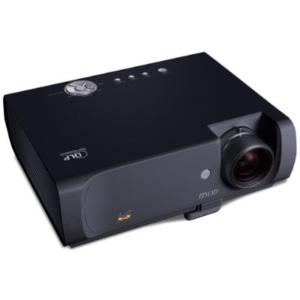}%
        \includegraphics[width=0.5\textwidth]{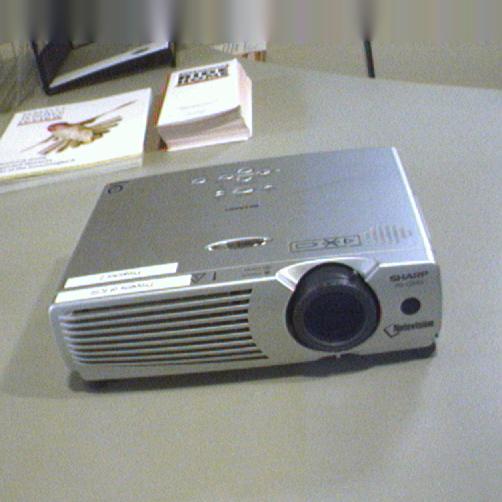}
	\end{minipage}} & \bmvaHangBox{
	\begin{minipage}[c][1\width]{0.21\textwidth}
	   \centering
        \includegraphics[width=0.5\textwidth]{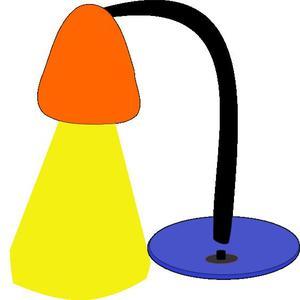}%
        \includegraphics[width=0.5\textwidth]{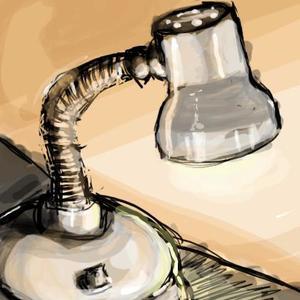}\\
        \includegraphics[width=0.5\textwidth]{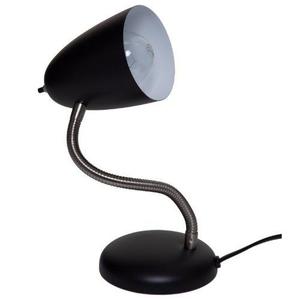}%
        \includegraphics[width=0.5\textwidth]{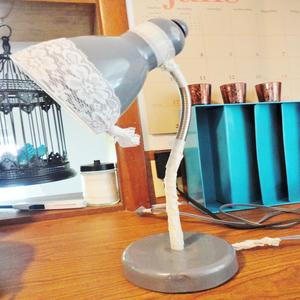}\\
	\end{minipage}}\\
	(a) VisDA 2017 & (b) Office-Caltech & (c) Office-31 & (d) Office-Home\\
\end{tabular}
\caption{From left to right: VisDA 2017~\cite{visda2017} with domains synthetic (train set) and real (validation and test set), Office-Caltech~\cite{officecaltech} with domains Amazon, Caltech, DSLR and Webcam, Office-31~\cite{office31} with domains Amazon, DSLR and Webcam and Office-Home~\cite{officehome} with domains Art, Clipart, Product and Real World.}
\label{fig:datasets}
\end{figure}

\subsection{Training Loop} \label{section:training_loop}

With the major parts of our training pipeline described above, we provide an overview for the UBR$^2$S training process in Figure~\ref{fig:decision_error}b. After the uncertainty extraction and resampling process, $\beta$ classes are sampled from class set $\mathcal{C}$. For every class $c$, $\frac{\lvert b \rvert}{2 \beta}$ samples are randomly drawn from the source and target bins $\mathcal{B}_c^{\mathcal{S}}$ and $\mathcal{B}_c^{\mathcal{T}}$ where $\lvert b \rvert$ is the batch size. The source bins are constructed based on the available ground truth labels while the target bins are reconstructed based on the label estimation described in Section~\ref{section:resampling_strat}. 

Finally, we calculate the sample likelihood and decision error for all target instances in a mini-batch and use it to reweigh their element-wise loss during training. Given the k-th target example $x_k$ in a mini-batch with a total of $K$ target instances we compute our proposed reweighted cross entropy loss as Equation~\ref{eq:target_loss} with weight $\omega$, where $v(\cdot)$ represents the label smoothing function from Section~\ref{section:dss} when using DSS and the one-hot encoding function otherwise.

\begin{equation}
    \mathcal{L}(x_k, \Tilde{y}_k)= -\omega_k \sum_{c \in \mathcal{C}} v(\Tilde{y}_k)_c\, \mathrm{log}\left[ g\left(f\left(x_k\right)\right)_c \right],\; \mathrm{with}\;\; \omega_k = \frac{\lambda_\mathrm{DE}^k\lambda_\mathrm{SL}^k}{\frac{1}{K} \sum_j^K \lambda_\mathrm{DE}^j\lambda_\mathrm{SL}^j} \label{eq:target_loss}
\end{equation}
\newline
 For source domain instances, weight $\omega$ is set to 1. Parameters $\theta$ of $f$ and $g$ are then updated by backpropagation according to this loss. Subsequent iterations use the updated weights $\theta$ for the uncertainty extraction and training process. Therefore, only a single neural network is needed during the complete training and adaptation phase.

\section{Experiments}
\textbf{Datasets}. We evaluate our proposed UBR$^2$S method on four public benchmark datasets: 
\textit{VisDA 2017} (also known as Syn2Real-C) \cite{visda2017} is a large scale dataset for the synthetic to real UDA task. It contains 12 classes in three domains: train (152,397 synthetic 3D renderings), validation (55,388 real world images from MS COCO~\cite{mscoco}) and test (72,372 real world images from YouTube Bounding-Boxes~\cite{real2017youtube}).
For comparison to state-of-the-art methods, we calculate the mean class accuracy \wrt to the challenge evaluation protocol unless otherwise noted.
\textit{Office-31}~\cite{office31} is one of the most used UDA datasets and contains 4,110 images of 31 classes in a generic office setting. The images come from the three domains Amazon (product images), DSLR and Webcam.
The \textit{Office-Caltech}~\cite{officecaltech} dataset is constructed from the 10 overlapping classes between the Caltech-256~\cite{caltech} and Office-31~\cite{office31} datasets for a total of four domains: Amazon (958), Caltech (1,123), DSLR (157) and Webcam (295).
The \textit{Office-Home}~\cite{officehome} dataset offers 65 challenging classes from everyday office life. Its four domains Art, Clipart, Product and Real World contain a total of 15,588 images. Example images from all datasets are shown in Figure~\ref{fig:datasets}. 
\newline
\textbf{Setup}. For our experiments, we follow the standard unsupervised domain adaptation setup (see \cite{can,dsbn}) and use all labeled source domain and all unlabeled target domain images for training. For a detailed description of the employed setup and training procedure, please refer to Appendix~\ref{sec:implementation}.

\begin{table}[t]
\centering
\resizebox{0.7\textwidth}{!}{
\begin{tabular}{cccrrrrrrrr}
\toprule
\rotatebox{0}{DSS\textsubscript{Pre}} & \rotatebox{0}{DSS\textsubscript{Ada}} & \rotatebox{0}{\small Reweigh} &
\rotatebox{0}{S$\underset{\text{Pre}}{\rightarrow}$R\textsubscript{test}} & \rotatebox{0}{S$\underset{\text{}}{\rightarrow}$R\textsubscript{test}} & \rotatebox{0}{Ar$\underset{\text{Pre}}{\rightarrow}$Cl} &\rotatebox{0}{Ar$\underset{\text{}}{\rightarrow}$Cl} & \rotatebox{0}{Pr$\underset{\text{Pre}}{\rightarrow}$Ar} & \rotatebox{0}{Pr$\underset{\text{}}{\rightarrow}$Ar}\\\toprule
$\times$ & $\times$ & $\times$ & 49.4 & 78.7 & 44.0 & 52.5 & 51.1 & 58.1\\
$\mathcal{S}$ & $\times$ & $\times$ & \textbf{51.3} & 82.5 & \textbf{46.0} & 54.1 & \textbf{54.1} & 58.8 \\
$\mathcal{S}$ & $\mathcal{S}$ & $\times$ & \textbf{51.3}  & 83.8 & \textbf{46.0} & 54.6 & \textbf{54.1} & 61.8\\
$\mathcal{S}$ & $\mathcal{T}$ & $\times$ &  \textbf{51.3}  & 67.0 & \textbf{46.0} & 38.6 & \textbf{54.1} & 47.5\\
$\mathcal{S}$ & $\mathcal{S},\mathcal{T}$ & $\times$ & \textbf{51.3}  & 67.8 & \textbf{46.0} & 47.4 & \textbf{54.1} & 53.6\\
$\mathcal{S}$ & $\mathcal{S}$ & SL & \textbf{51.3}  & 85.4 & \textbf{46.0} & 56.7 & \textbf{54.1} & 64.1\\
$\mathcal{S}$ & $\mathcal{S}$ & DE & \textbf{51.3}  & 89.5 & \textbf{46.0} & 57.4 & \textbf{54.1} & 65.1\\\hline\Tstrut
$\mathcal{S}$ & $\mathcal{S}$ & DE+SL & \textbf{51.3} & \textbf{89.8} &\textbf{ 46.0} & \textbf{58.3} & \textbf{54.1} & \textbf{67.0}\\
\toprule
\end{tabular}}

\caption{Ablation study on the VisDA 2017 S$\rightarrow$R\textsubscript{test} (mean class accuracy) and Office-Home Ar$\rightarrow$Cl, Pr$\rightarrow$Ar tasks (accuracy). Transfer tasks marked with \raisebox{0.4em}{$\underset{\text{Pre}}{\rightarrow}$} indicate results after the source only pretraining and before the adaptation step. The last table row represents our full UBR$^2$S method. \label{table:ablation}}
\end{table}

\subsection{Results}
\textbf{Ablation Study}. We first conduct an ablation study \wrt to every part of our proposed UBR$^2$S method. For this, we use ResNet-101 on the VisDA 2017 S$\rightarrow$R (test set) task as well as ResNet-50 on the Ar$\rightarrow$Cl and Pr$\rightarrow$Ar transfer tasks from Office-Home. Results are reported in Table~\ref{table:ablation}.
First, we examine our proposed domain specific smoothing (DSS) method. Our baseline does not use DSS during pretraining (DSS$_\text{Pre}^\times$) or during the adaptation phase (DSS$_\text{Ada}^\times$). Expectedly, this baseline performs the worst for all three transfer tasks due to overfitting on the source domain. Applying DSS to the source samples during the pretraining phase (DSS$_\text{Pre}^\mathcal{S}$) already improves pretrained results by almost 2\% for VisDA (49.4\% to 51.3\%) and final results by almost 4\% (78.7\% to 82.5\%). Similar trends can be observed for the Office-Home transfer tasks.
Concerning DSS$_\text{Ada}$, we evaluate all possible combinations $\{\times, \mathcal{S}, \mathcal{T}, \mathcal{S}+\mathcal{T}\}$. We find that applying label smoothing to the target domain (DSS$_\text{Pre}^\mathcal{T}$ and DSS$_\text{Pre}^{\mathcal{S},\mathcal{T}}$) has a negative impact on the model's accuracy. This is consistent for all three transfer tasks and can reduce accuracy by up to 16.8\%. Conversely, adding label smoothing to the source domain always improves performance, considering both the DSS$_\text{Ada}^\times$ to DSS$_\text{Ada}^\mathcal{S}$ and DSS$_\text{Ada}^\mathcal{T}$ to DSS$_\text{Ada}^{\mathcal{S},\mathcal{T}}$ transitions.
Overall, DSS$_\text{Pre}^\mathcal{S}$ with DSS$_\text{Ada}^\mathcal{S}$ consistently achieved the best results. This confirms our hypothesis from Section~\ref{section:dss} and implies that a non-saturated source classifier is needed for good transferability and generalization to new domains in UDA tasks. Further ablation studies are shown in the appendices.

We continue by examining the remaining parts of our UBR$^2$S method in Table~\ref{table:ablation} using the best performing DSS$_\text{Pre}^\mathcal{S}$ with DSS$_\text{Ada}^\mathcal{S}$ setup as baseline. As mentioned in Section~\ref{sec:reweighting_strat}, the resampling process in UBR$^2$S comes with an inherent risk and the possibility for errors. This risk can be partially measured with the help of our proposed sample likelihood (SL) and used for reweighting, which already improves results by 1.6\% for VisDA and up to 2.3\% for the Office-Home tasks. It is also important to assess the current label estimation and how the chosen pseudo-label compares to other potential candidates. This is covered by our proposed decision error (DE), which -- when solely used for reweighting -- can improve results by 5.7\% for VisDA and up to 3.3\% for the Office-Home tasks. Collectively, the combination of SL and DE can further improve results by 6.0\% for VisDA and up to 5.2\% for Office-Home over the respective baselines and constitutes our full UBR$^2$S method.

\begin{table*}[t]
\centering
\resizebox{1.0\textwidth}{!}{
\begin{tabular}{l@{\extracolsep{4pt}}rrrrrrrrrrrrr}
\toprule
    \multirow{2}{*}{Method} & \multicolumn{3}{c}{A} & \multicolumn{3}{c}{C} & \multicolumn{3}{c}{D} & \multicolumn{3}{c}{W} & \multirow{2}{*}{Avg.}\\\cline{2-4}\cline{5-7}\cline{8-10}\cline{11-13}
     & C & D & W & A & D & W & A & C & W & A & C & D & \\\toprule
    CORAL \cite{sun2016return,rwa} & 89.2 & 92.2 &  91.9 &  94.1 &  92.0 &  92.1  & 94.3 &  87.7 &  98.0 &  92.8 &  86.7 & \textbf{100.0} & 92.6\\
    GTDA+LR \cite{vascon2019unsupervised}  & 91.5 & 98.7 & 94.2 & 95.4 & 98.7 & 89.8 & 95.2 & 89.0 & 99.3 & 95.2 & 90.4 & {\textbf{100.0}} & 94.8\\
    RWA \cite{rwa} & 93.8 & {98.9} & {97.8} & 95.3 & \underline{99.4} & 95.9 & 95.8 & 93.1 & 98.4 & 95.3 & 92.4 & \underline{99.2} & 96.3\\
    PrDA \cite{hua2020unsupervised} & 92.1 & 99.0 & \underline{99.3} & \textbf{97.2} & \underline{99.4} & 98.3 & 94.7 & 91.0 & \underline{99.7} & 95.6 & 93.4 & \textbf{100.0} & 96.6 \\
    Rakshit \etal \cite{ccduda}$\ast$ & 92.8 & {98.9} & 97.0 & {96.0} & {99.0} & 97.0  & \underline{96.5}  & {\textbf{97.0}}  & 99.5  & 95.5  & 91.5  & {\textbf{100.0}} & {96.8}\\
    ACDA \cite{zhang2021adversarial} & \underline{93.9} & \textbf{100.0} & \textbf{100.0} & 96.2 & \textbf{100.0} & \textbf{100.0} & \textbf{96.7} & \underline{93.9} & \textbf{100.0} & \textbf{96.6} & 93.9 & \textbf{100.0} & \textbf{97.6} \\\hline\Tstrut
    \textbf{UBR$^2$S (ours)} & \textbf{95.5} & \underline{99.4} & \underline{99.3} & \underline{96.6} & 94.9 & \underline{99.7} & 96.2 & \underline{95.3} & \textbf{100.0} & \underline{96.2} & \textbf{95.4} & \textbf{100.0} & \underline{97.4} \\
\toprule
\end{tabular}}
\caption{\label{table:officecaltech_comparison}Classification accuracy (in \%) for different methods using ResNet-50 on the \textbf{Office-Caltech} dataset with domains Amazon, Caltech, DSLR and Webcam. The method marked with $\ast$ uses an ensemble setup with multiple classifiers.}
\end{table*}

\begin{table*}[t]
\centering
\begin{adjustbox}{width=1.\textwidth}
\begin{tabular}{lrrrrrrrrrrrrr}
\toprule
    Method & {aero} & {bicyc} & {bus} & {car} & {horse} & {knife} & {motor} & {person} & {plant} & {skate} & {train} & {truck} & {Avg.}\\\toprule
    Source only & 52.8 & 13.8 & 66.9 & 96.3 & 58.4 & 14.0 & 63.4 & 34.5 & 86.0 & 24.5 & 87.3 & 17.9 & 51.3\\
    BUPT \cite{visda2017} & \underline{95.7} & 67.0 & {93.4} & \textbf{97.2} & {90.6} & 86.9 & \textbf{92.0} & 74.2 & \underline{96.3} & 66.9 & \textbf{95.2} & 69.2 & 85.4\\
    CAN \cite{can} & --- & --- & --- & --- & --- & --- & --- & --- & --- & --- & --- & --- & 87.4 \\
    SDAN~\cite{sdan} & 94.3 & {86.5} & {86.9} & {95.1} & {91.1} & {90.0} & \underline{82.1} & {77.9} & {96.4} & {77.2} & 86.6 & \underline{88.0} & {87.7}\\
    UFAL~\cite{ringwald2020unsupervised} & 94.9 & \underline{87.0} & \underline{87.0} & \underline{96.5} & \underline{91.8} & \textbf{95.1} & 76.8 &\textbf{78.9} & \textbf{96.5} & \underline{80.7} & \underline{93.6}& 86.5 & \underline{88.8}\\\hline\Tstrut
    \textbf{UBR$^2$S (ours)} &\textbf{ 96.6} &\textbf{ 90.8} & \textbf{87.9} & 94.6 & \textbf{92.2} & \underline{92.8} & 77.8 & \underline{78.8} & 95.3 &\textbf{89.2} & 92.6 & \textbf{88.9} &\textbf{89.8}\\
\toprule
\end{tabular}
\end{adjustbox}
\caption{\label{table:visda_test_sota}Per class accuracy (in \%) for different methods on the \textbf{VisDA 2017 test set} as per challenge evaluation protocol. Results are obtained using the common ResNet-101 backbone.}
\end{table*}

\textbf{Comparison to state-of-the-art}. We continue by comparing UBR$^2$S to other recently proposed approaches. Results for the Office-Caltech dataset are shown in Table~\ref{table:officecaltech_comparison}. Evidently, UBR$^2$S can achieve domain adaptation even for Office-Caltech's 12 diverse transfer tasks. Our proposed method achieves the best or second best result in 10 out of 12 transfer tasks (such as A$\rightarrow$C) and is also on par with ACDA~\cite{zhang2021adversarial} for the overall average. Notably, UBR$^2$S even manages to surpass the ensemble-based setup of Rakshit~\etal~\cite{ccduda} by 0.6\%.
Additionally, we report results for the VisDA 2017 test set in Table~\ref{table:visda_test_sota} and calculate the class accuracies as per challenge evaluation protocol. Our results indicate that UBR$^2$S can also achieve domain adaptation for VisDA's difficult synthetic to real transfer task and outperforms recently proposed methods. With 89.8\% mean class accuracy, UBR$^2$S also outperforms the VisDA challenge submissions SDAN~\cite{sdan} and BUPT~\cite{visda2017}. Given the current VisDA 2017 challenge leaderboard~\cite{visdaleaderboard}, UBR$^2$S would rank second place -- only behind SE~\cite{french2017self}, a 5$\times$ResNet-152 ensemble with results averaged over 16 test time augmentation runs. This, however, is not a fair comparison to our single ResNet-101 model, but still demonstrates UBR$^2$S' competitive UDA capabilities.

\begin{figure}
    \centering
    \includegraphics[width=0.48\columnwidth]{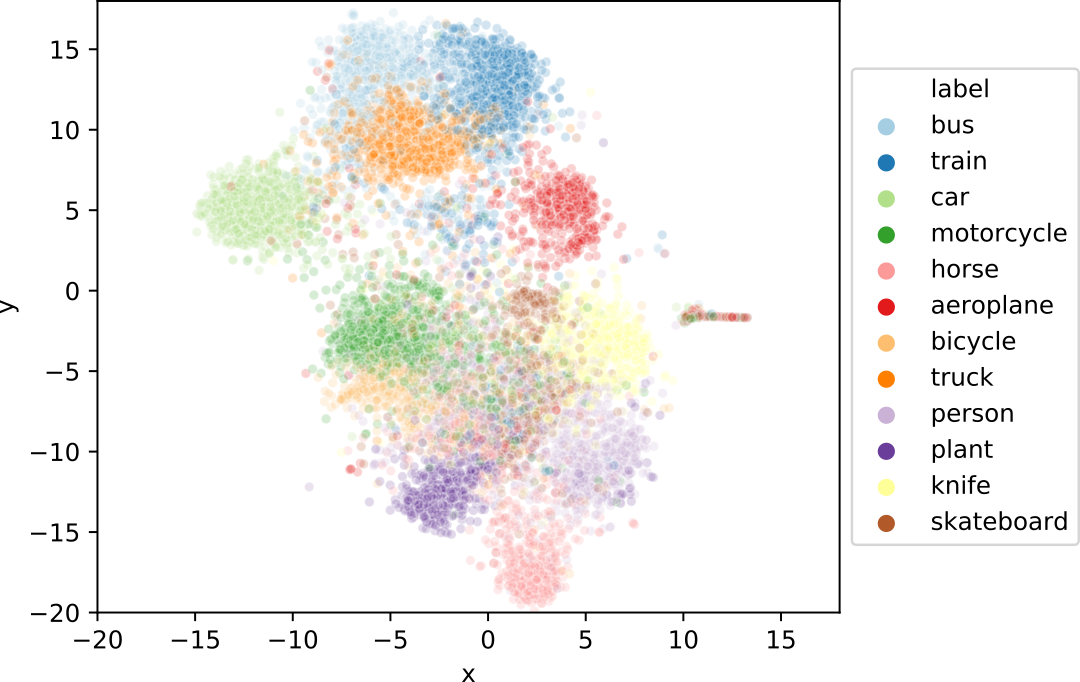}\hfill
    \includegraphics[width=0.48\columnwidth]{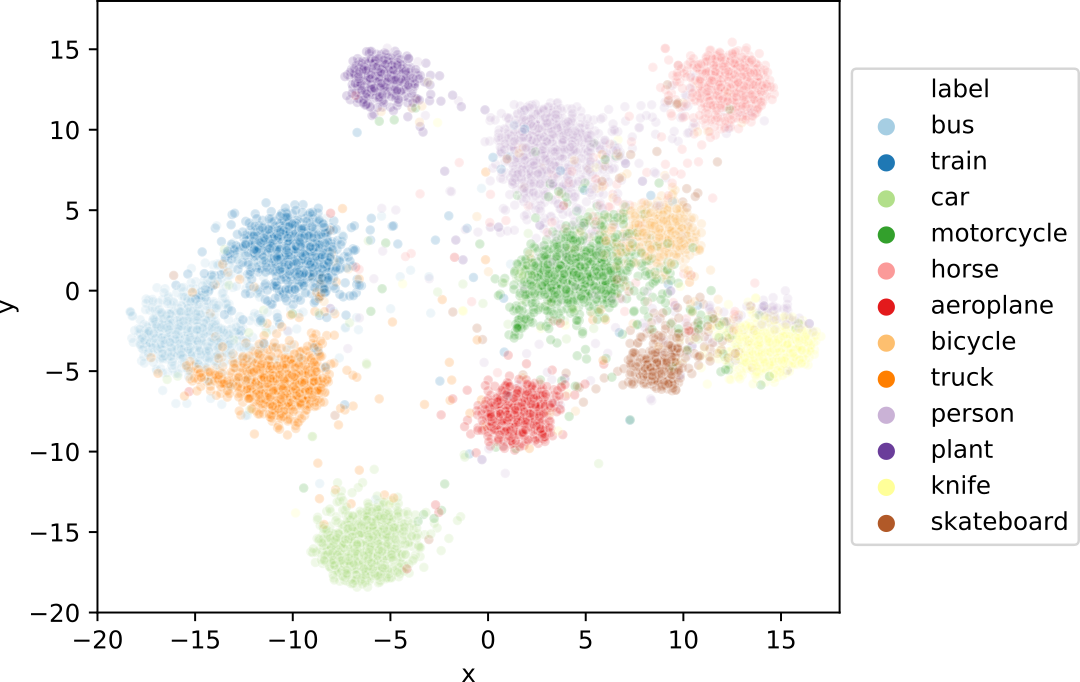}
    \caption{Visualizations of the VisDA 2017 test set features using t-SNE. Left: After the source-only pretraining phase. Right: After the adaptation phase using UBR$^2$S.}
    \label{fig:tsneplot}
\end{figure}

\textbf{Visualization}
Finally, we visualize the embeddings learned by UBR$^2$S in Figure~\ref{fig:tsneplot}. For this, we extract target domain features for the VisDA 2017 test set (real domain) before and after the adaptation phase and project them into 2D space via t-SNE~\cite{maaten2008visualizing}. 
After the source-only pretraining phase, the model clearly has not learned discriminative representations for target domain samples. Features of all classes are accumulated in one big cluster with no clear separation. After UBR$^2$S' unsupervised domain adaptation phase, visually distinct clusters for each of VisDA's 12 classes can be observed. This indicates that UBR$^2$S is also able to learn discriminative target domain representations even in the absence of target domain annotations.

We provide further results in the appendices. This includes additional ablation studies, full results on the Office-Home~\cite{officehome} dataset, multi-source UDA results on the Office-31~\cite{office31} dataset, an evaluation with different network backbones and an analysis of training stability.

\section{Conclusion}

In this paper, we propose UBR$^2$S, the uncertainty-based resampling and reweighting strategy. UBR$^2$S' resampling phase is based on a model's prediction uncertainty quantified by Monte Carlo dropout. As resampling introduces the possibility of sampling a wrong pseudo-label, a dynamic reweighting stage is added to assess and incorporate this risk in the loss calculation. The efficacy of UBR$^2$S is shown on multiple UDA benchmark datasets such as VisDA 2017, Office-Caltech, Office-31 and Office-Home in single and multi-source domain adaptation setups in which UBR$^2$S outperforms recently proposed methods and achieves state-of-the-art results. Furthermore, we show that UBR$^2$S can be applied to any off-the-shelf CNN and works even with very small networks (such as MobileNetV2) with extremely low parameter counts. Our code is made available on the project website for reproduction of our results and to encourage further research in the area of unsupervised domain adaptation.

\clearpage
\setcounter{section}{0}
\setcounter{table}{0}
\setcounter{figure}{0}

\renewcommand{\thetable}{\Roman{table}}
\renewcommand{\thefigure}{\Roman{figure}}

\appendix
\section*{Appendices}
\addcontentsline{toc}{section}{Appendices}
\renewcommand{\thesubsection}{\Alph{subsection}}

\subsection{Overview}

We provide further experimental results in the following sections. Section~\ref{sec:implementation} describes the employed hyperparameters, network architectures and training setup. Section~\ref{sec:ablations} provides ablation studies for the $\varepsilon$ hyperparameter and training stability over multiple runs. Section~\ref{sec:multisource} reports results on the Office-31~\cite{office31} and Office-Home~\cite{officehome} datasets in a multi-source UDA setup. Section~\ref{sec:architectures} studies UBR$^2$S' performance \wrt different backbone architectures.

Furthermore, we provide additional experimental results on Office-Home~\cite{officehome} and the VisDA 2017~\cite{visda2017} validation set in Tables~\ref{table:officehome_comparison} and~\ref{table:visda_sota}. An expanded version of the ablation study in the main paper can be found in Table~\ref{table:ablation_full}.

\subsection{Implementation Details}\label{sec:implementation}
\textbf{Hyperparameters}. All of our experiments are based on the same hyperparameter set and follow the setup proposed in \cite{ringwald2020unsupervised}: We first optimize $f$ and $g$ on the source domain for 1000 iterations using SGD with batch size 240 and learning rate $5\times10^{-4}$. For the adaptation phase, the learning rate is $2.5\times10^{-4}$ over 100 cycles with 50 forward passes each cycle and $\beta$=$12$ (as per \cite{ringwald2020unsupervised}). The Monte-Carlo dropout rate is 75\% with $\lvert\mathcal{M}\rvert$=50 for all setups. Experiments involving DSS use $\varepsilon$=$0.25$ (see Section~\ref{sec:ablations}). 
Our method is implemented in PyTorch~\cite{paszke2017automatic} and trained on four NVIDIA 1080 Ti GPUs.
\newline
\textbf{Network architectures}. For Office-Caltech, Office-31 and Office-Home, we utilize ResNet-50~\cite{he2016deep} for comparison to SOTA. For our VisDA 2017 experiments, we use the default ResNet-101~\cite{he2016deep} architecture unless otherwise noted. During the backbone ablation study, we also employ MobileNetV2~\cite{sandler2018mobilenetv2} and DenseNet-121~\cite{huang2017densely}. In any case, all networks are pretrained on ImageNet~\cite{imagenet}, use a two layer classifier $g$ (similar to \cite{roitberg2018informed}) and the loss described in the main paper.
\newline
\textbf{Multi-source bins}. In multi-source UDA setups with $D$ domains, source bins $\mathcal{B}_c^{\mathcal{S}_{1..D}}$ (see main paper) are created per-domain. At the start of every cycle, a domain $d$ is chosen from the available source domains, whose bin $\mathcal{B}_c^{\mathcal{S}_d}$ is then used for the construction of mini-batches.

\subsection{Ablation Studies}\label{sec:ablations}
\subsubsection{Hyperparameter $\varepsilon$}
\begin{figure}
    \centering
    \includegraphics[width=0.4\textwidth]{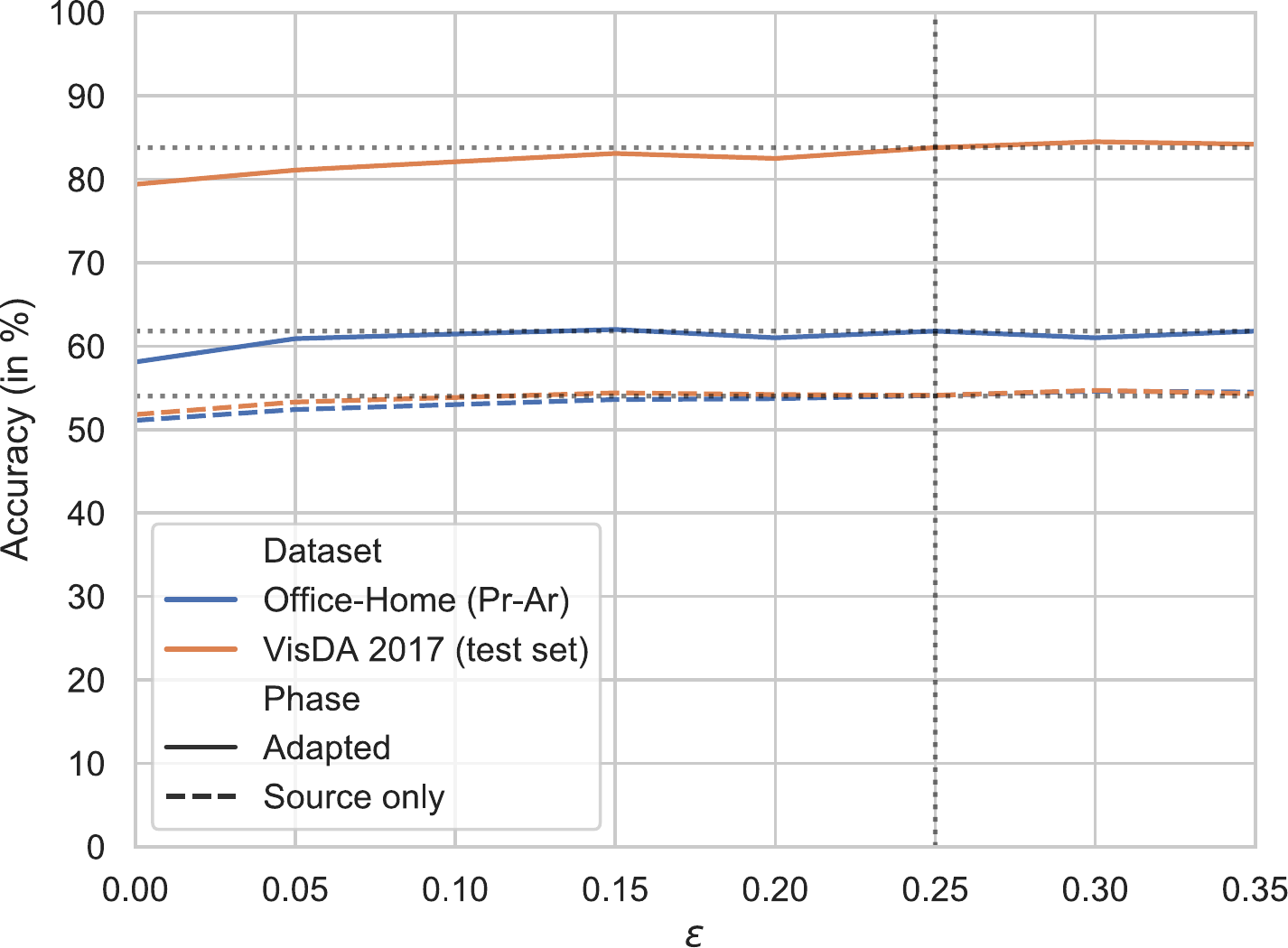}
    \caption{Ablation study for the $\varepsilon$ parameter of our proposed domain specific smoothing (DSS). Results are shown for the VisDA 2017 train$\rightarrow$test task and Office-Home's Pr$\rightarrow$Ar task before (source only) and after the adaptation step.}
    \label{fig:eps_ablation_visda}
\end{figure}

In Figure~\ref{fig:eps_ablation_visda}, we provide an ablation study for the $\varepsilon$ parameter of our proposed domain specific smoothing (DSS) setup. Results are generated on the VisDA 2017~\cite{visda2017} test set and Office-Home's~\cite{officehome} Pr$\rightarrow$Ar task after the source only pretraining phase and after the adaptation step (using the DSS$_\text{Pre}^\mathcal{S}$, DSS$_\text{Ada}^\mathcal{S}$ setup from the main paper). Setting $\varepsilon=0$ is equivalent to not using label smoothing at all (DSS$_\text{Pre}^\times$, DSS$_\text{Ada}^\times$) and thus yields the worst results. In line with our hypothesis in the main paper, using label smoothing increases the domain adaptation capabilities. We notice that the results are robust for $\varepsilon \geq 0.15$ considering both source only pretraining and the adapted results. Overall, setting $\varepsilon = 0.25$ yields the most consistent performance for the examined transfer tasks and was hence chosen as basis for our further experiments.

\subsubsection{Training Stability}
In Table~\ref{table:visda_stability}, we additionally analyze UBR$^2$S' stability over multiple runs with consecutive random seeds. Results are reported for the VisDA 2017~\cite{visda2017} train$\rightarrow$val and train$\rightarrow$test transfer tasks and averaged over 3 runs. Expectedly, the source only pretraining exhibits larger fluctuations as the model has not been trained on any target domain data at this point. However, after the adaptation step with UBR$^2$S, results consistently converge towards similar values with very minor deviations of $\pm 0.3$\% and $\pm 0.1$\% accuracy for the validation and test set respectively. Similar results can also be noted for the VisDA challenge evaluation protocol metric (mean class accuracy). We thus conclude that our proposed UBR$^2$S method is also stable over multiple runs with different random seeds.

\subsection{Multi-source DA}\label{sec:multisource}
We also evaluate UBR$^2$S in a multi-source domain adaptation setting and report results in Table~\ref{table:multisource}. For this, we employ the setup of~\cite{dsbn,zhu2019aligning} and report results for Office-31 (D,W$\rightarrow$A) and Office-Home (Ar,Cl,Pr$\rightarrow$Rw) in two settings: \emph{Source combine} merges all available source domains into a single, larger source domain while the \emph{multi-source} setup keeps information about the source domain affiliation of all source samples available during training.
For Office-31, UBR$^2$S is able to surpass DSBN~\cite{dsbn} by 4.4\% in the source combine setting and 1.7\% in the multi-source setting. 
For Office-Home, UBR$^2$S is able to outperform MFSAN~\cite{zhu2019aligning} by 0.5\% and DSBN~\cite{dsbn} by 0.9\% in the source combine setup. UBR$^2$S does also exceed MFSAN~\cite{zhu2019aligning} by 1.7\% and the very recent WAMDA~\cite{wamda} method by 1.2\% in the multi-source setup.
These results imply that our proposed UBR$^2$S method also has an advantage in multi-source domain adaptation settings and is able to learn from multiple different source domains at once.

\subsection{Backbone Architectures}\label{sec:architectures}
One of the key advantages of UBR$^2$S is that it can be applied to any off-the-shelf network. No additional layers or auxiliary networks (\eg generators or discriminators) are required. We therefore also provide results for different network architectures to show UBR$^2$S' performance \wrt parameter count. We choose ResNet-50~\cite{he2016deep}, ResNet-101~\cite{he2016deep}, DenseNet-121~\cite{huang2017densely} and MobileNetV2~\cite{sandler2018mobilenetv2} for comparison and evaluate on the VisDA 2017 validation set. Parameter counts and results are depicted in Table~\ref{table:visda_backbone} as both accuracy and mean class accuracy as per challenge evaluation protocol.

Most commonly, results are reported using ResNet-50 and ResNet-101. In these categories, UBR$^2$S is able to outperform very recently proposed methods such as STAR~\cite{lu2020stochastic}, UFAL~\cite{ringwald2020unsupervised} and the approach proposed by Li~\etal~\cite{li2020model}.
When comparing ResNet-50 to ResNet-101 results, we note that UBR$^2$S' accuracy only drops by 3.2\%, even though ResNet-50 has approximately 19 million fewer parameters. Additionally, even at this reduced parameter count, UBR$^2$S is able to outperform the larger ResNet-152 results of SimNet~\cite{simnet} and GTA~\cite{sankaranarayanan2018generate} by a large margin.

For MobileNetV2 and DenseNet-121, no comparative values exist in literature. However, it is noteworthy that UBR$^2$S is still able to perform on par with ResNet-50 results at one third (DenseNet-121) and one tenth (MobileNetV2) of their parameters.
We thus show that UBR$^2$S is indeed applicable to arbitrary off-the-shelf network architectures. This opens up opportunities for faster research prototyping, lower resource requirements while training and the option for an accuracy-speed trade-off in deployment scenarios.

\begin{table}[h!]
    \centering
    \begin{adjustbox}{width=0.7\columnwidth}
    \begin{tabular}{l@{\extracolsep{4pt}}cccc}\toprule
        \multirow{2}{*}{Method} & \multicolumn{2}{c}{Source Combine} & \multicolumn{2}{c}{Multi-Source}  \\\cline{2-3}\cline{4-5}
         & D,W$\rightarrow$A & Ar,Cl,Pr$\rightarrow$Rw & D,W$\rightarrow$A & Ar,Cl,Pr$\rightarrow$Rw\\\toprule
         BN~\cite{dsbn} & 71.3 & 81.2 & 69.9 & 81.4 \\
         WAMDA~\cite{wamda} & --- & --- & 72.0 & 82.3 \\
         MFSAN~\cite{zhu2019aligning} & 67.6 & 82.7 & 72.7 & 81.8 \\
         DSBN~\cite{dsbn} & 73.2 & 82.3 & 75.6 & 83.0 \\\hline\Tstrut
         \textbf{UBR$^2$S (ours)} & \textbf{77.6} & \textbf{83.2} & \textbf{77.3} & \textbf{83.5} \\\toprule
    \end{tabular}
    \end{adjustbox}
    \caption{Classification accuracy (in \%) for different methods when leveraging multiple source domains of the Office-31 (D, W) and Office-Home (Ar, Cl, Pr) datasets.}
    \label{table:multisource}
\end{table}

\begin{table}[h!]
\centering
    \begin{adjustbox}{width=.5\columnwidth}
        \begin{tabular}{lclrr}
        \toprule
        Backbone & Parameters & Method & Acc. & Mean Acc.\\\toprule
        MobileNetV2 & 2.2$\times 10^6$ & \textbf{UBR$^2$S (ours)} & \textbf{70.1} & \textbf{69.5} \\\hline\Tstrut
        \vphantom{$\sum^T$}DenseNet-121 & 7.0$\times 10^6$ & \textbf{UBR$^2$S (ours)} & \textbf{77.6} & \textbf{79.5} \\\hline\Tstrut
        \multirow{6}{*}{ResNet-50} & \multirow{6}{*}{23.5$\times 10^6$} & GTA \cite{sankaranarayanan2018generate} & 69.5 & --- \\
        & & SimNet \cite{simnet} & --- & 69.6 \\
        & & CDAN+E \cite{long2018conditional} & 70.0 & --- \\
        & & TAT \cite{liu2019transferable} & 71.9 & --- \\
        & & DTA \cite{lee2019drop} & --- & 76.2 \\
        & & \textbf{UBR$^2$S (ours)} & \textbf{79.0} & \textbf{79.8} \\\hline\Tstrut
        \multirow{6}{*}{ResNet-101} & \multirow{6}{*}{42.5$\times 10^6$} & DSBN \cite{dsbn} & --- & 80.2 \\
        & & DTA \cite{lee2019drop} & --- & 81.5 \\
        & & STAR \cite{lu2020stochastic} & --- & 82.7 \\
        & & Li~\etal~\cite{li2020model} & --- & 83.3 \\
        & & UFAL \cite{ringwald2020unsupervised} & 81.8 & 84.7 \\
        & & \textbf{UBR$^2$S (ours)} &\textbf{ 82.2} & \textbf{85.2} \\\hline\Tstrut
        \multirow{2}{*}{ResNet-152} & \multirow{2}{*}{58.1$\times 10^6$} & SimNet \cite{simnet} & --- & \textbf{72.9} \\
        & & GTA \cite{sankaranarayanan2018generate} & \textbf{77.1} & --- \\
        \toprule
        \end{tabular}
    \end{adjustbox}
\caption{Classification accuracy and mean class accuracy (in \%) for different network architectures and methods on the VisDA 2017 validation set.}\label{table:visda_backbone}
\end{table}

\begin{table}[h]
\centering
\resizebox{\textwidth}{!}{
\begin{tabular}{llrrrrrrrrrrrrrr}
\toprule
    Subset & Method & {\small aero} & {\small bicyc} & {\small bus} & {\small car} & {\small horse} & {\small knife} & {\small motor} & {\small person} & {\small plant} & {\small skate} & {\small train} & {\small truck} & {Mean Acc.} & Accuracy\\\toprule
    Train$\rightarrow$Val & Source only & \z{59.8}{10.9} & \z{19.8}{7.7} & \z{63.4}{3.8} & \z{72.9}{3.3} & \z{73.8}{2.9} & \z{15.6}{2.3} & \z{81.5}{5.1} & \z{39.8}{8.4} & \z{71.7}{3.8} & \z{31.2}{2.5} & \z{87.7}{0.8} & \z{7.8}{0.6} & \z{52.1}{2.4} & \z{57.1}{1.9}\\
    Train$\rightarrow$Val & UBR$^2$S & \z{97.6}{0.2} & \z{83.3}{1.0} & \z{81.7}{1.9} & \z{70.9}{2.7} & \z{95.7}{0.4} & \z{93.3}{3.1} & \z{89.1}{0.6} & \z{84.3}{1.5} & \z{94.8}{1.1} & \z{91.4}{1.5} & \z{89.0}{0.8} & \z{52.8}{2.5} & \textbf{\z{85.3}{0.2}} & \textbf{\z{82.5}{0.3}}\\\hline\Tstrut
    Train$\rightarrow$Test & Source only & \z{54.4}{10.3} & \z{10.4}{4.4} & \z{70.6}{3.3} & \z{95.3}{1.0} & \z{61.0}{2.5} & \z{15.7}{3.3} & \z{71.3}{7.0} & \z{27.4}{7.8} & \z{86.0}{2.3} & \z{29.5}{4.4} & \z{84.7}{2.3} & \z{18.3}{0.4} & \z{52.1}{1.7} & \z{54.7}{1.6} \\
    Train$\rightarrow$Test & UBR$^2$S & \z{96.6}{0.1} & \z{88.5}{2.1} & \z{88.3}{0.8} & \z{95.2}{0.6} & \z{92.4}{1.2} & \z{95.1}{2.3} & \z{77.5}{0.3} & \z{78.9}{0.4} & \z{96.4}{1.1} & \z{85.4}{3.9} & \z{93.0}{1.2} & \z{87.0}{2.0} & \textbf{\z{89.5}{0.3}} & \textbf{\z{89.0}{0.1}}\\ 
\toprule
\end{tabular}}
\caption{\label{table:visda_stability} Stability analysis for our final UBR$^2$S method on the VisDA 2017 validation and test set averaged over 3 runs with consecutive random seeds. Results are obtained with a ResNet-101 backbone.}
\end{table}

\begin{table}[h]
\centering
\begin{adjustbox}{width=1.\textwidth}
\begin{tabular}{l@{\extracolsep{4pt}}rrrrrrrrrrrrr}
\toprule
    \multirow{2}{*}{Method} & \multicolumn{3}{c}{Ar} & \multicolumn{3}{c}{Cl} & \multicolumn{3}{c}{Pr} & \multicolumn{3}{c}{Rw} & \multirow{2}{*}{Avg.}\\\cline{2-4}\cline{5-7}\cline{8-10}\cline{11-13}
     & Cl & Pr & Rw & Ar & Pr & Rw & Ar & Cl & Rw & Ar & Cl & Pr & \\\toprule
    TAT \cite{liu2019transferable} & 51.6 & 69.5 & 75.4 & 59.4 & 69.5 & 68.6 & 59.5 & 50.5 & 76.8 & 70.9 & 56.6 & 81.6 & 65.8\\
    ETD~\cite{li2020enhanced} & 51.3 & 71.9 & \textbf{ 85.7} & 57.6 &  69.2 & 73.7 & 57.8 & 51.2 & 79.3 & 70.2 & 57.5 & 82.1 & 67.3 \\
    MDDA \cite{wang2020transfer} & 54.9 & 75.9 & 77.2 & 58.1 & 73.3 & 71.5 & 59.0 & 52.6 & 77.8 & 67.9 & 57.6 & 81.8 & 67.3\\
    CDAN+TransNorm \cite{wang2019transferableNIPS} & 50.2 & 71.4 & 77.4 & 59.3 & 72.7 & 73.1 & 61.0 & 53.1 & 79.5 & 71.9 & 59.0 & 82.9 & 67.6\\
    CADA-P \cite{kurmi2019attending} & 56.9 & {\textbf{76.4}} & {80.7} & 61.3 & {\textbf{75.2}} & {\textbf{75.2}} & 63.2 & 54.5 & {\textbf{80.7}} & \textbf{{73.9}} & 61.5 & \textbf{{84.1}} & 70.2\\
    GSDA \cite{hu2020unsupervised} & 61.3 & 76.1 & 79.4 & 65.4 & 73.3 & 74.3 & 65.0 & 53.2 & 80.0 & 72.2 & 60.6 & 83.1 & 70.3\\
    UFAL \cite{ringwald2020unsupervised} & 58.5 & 75.4 & 77.8 & 65.2 & 74.7 & 75.0 & 64.9 & 58.0 & 79.9 & 71.6 & 62.3 & 81.0 & 70.4\\
    CAPLS \cite{wang2019unifying} & 56.2 & 78.3 & 80.2 & 66.0 & 75.4 & 78.4 & 66.4 & 53.2 & 81.1 & 71.6 & 56.1 & 84.3 & \textbf{70.6} \\\hline\Tstrut
    \textbf{UBR$^2$S (ours)} &\textbf{ 58.3} & 75.5 & 78.7 & \textbf{67.8} & 72.6 & 71.4 &\textbf{ 67.0} &\textbf{ 58.7} & 79.0 & 73.7 & \textbf{61.8} & 82.2 & \textbf{70.6}\\
\toprule
\end{tabular}
\end{adjustbox}
\caption{\label{table:officehome_comparison}Classification accuracy (in \%) for different methods on the Office-Home dataset with domains Art, Clipart, Product and Real-world.}
\end{table}

\begin{table}[h]
\centering
\resizebox{\textwidth}{!}{
\begin{tabular}{lrrrrrrrrrrrrr}
\toprule
    Method & {\small aero} & {\small bicyc} & {\small bus} & {\small car} & {\small horse} & {\small knife} & {\small motor} & {\small person} & {\small plant} & {\small skate} & {\small train} & {\small truck} & {Avg.}\\\toprule
    Source only & 59.1 & 23.3 & 59.1 & 76.1 & 73.8 & 15.7 & 75.7 & 47.0 & 70.5 & 31.6 & 88.5 & 7.3 & 52.3 \\
    SimNet-152 \cite{simnet} & 94.3 & 82.3 & 73.5 & 47.2 & 87.9 & 49.2 & 75.1 & 79.7 & 85.3 & 68.5 & 81.1 & 50.3 & 72.9 \\
    DSBN \cite{dsbn} & 94.7 & 86.7 & 76.0 & 72.0 & 95.2 & 75.1 & 87.9 & 81.3 & 91.1 & 68.9 & 88.3 & 45.5 & 80.2 \\
    DTA \cite{lee2019drop} & 93.7 & 82.8 & 85.6 & \textbf{83.8} & 93.0 & 81.0 & \textbf{90.7} & 82.1 & \textbf{95.1} & 78.1 & 86.4 & 32.1 & 81.5 \\
    STAR \cite{lu2020stochastic} & 95.0 & 84.0 & 84.6 & 73.0 & 91.6 & 91.8 & 85.9 & 78.4 & 94.4 & 84.7 & 87.0 & 42.2 & 82.7\\
    Li \etal \cite{li2020model} & 95.7 & 78.0 & 69.0 & 74.2 & 94.6 & 93.0 & 88.0 & \textbf{87.2} & 92.2 & 88.8 & 85.1 & 54.3 & 83.3\\
    SE-152 \cite{french2017self} & 95.9 & \textbf{87.4} & 85.2 & 58.6 & \textbf{96.2} & \textbf{95.7}& 90.6 & 80.0 & 94.8 & 90.8 & 88.4 & 47.9 & 84.3 \\
    UFAL \cite{ringwald2020unsupervised} & \textbf{97.6} & 82.4 & \textbf{86.6} & 67.3 & 95.4 & 90.5 & 89.5 & 82.0 & \textbf{95.1} & 88.5 & 86.9 & 54.0 & 84.7\\\hline\Tstrut
    \textbf{UBR$^2$S (ours)} & 97.5 & 84.4 & 81.5 & 67.9 & 95.5 & 89.8 & 88.7 & 85.8 & 93.6 & \textbf{93.0} & \textbf{89.4} & \textbf{55.6} & \textbf{85.2}\\
\toprule
\end{tabular}}
\caption{\label{table:visda_sota}Per class accuracy (in \%) for different methods on the VisDA 2017 validation set as per challenge evaluation protocol. Results are obtained with ResNet-101 unless otherwise denoted.}
\end{table}

\begin{table}[h]
\centering
\resizebox{\textwidth}{!}{
\begin{tabular}{cccrrrrrrrr}
\toprule
\rotatebox{0}{DSS\textsubscript{Pre}} & \rotatebox{0}{DSS\textsubscript{Ada}} & \rotatebox{0}{\small Reweigh} &
\rotatebox{0}{S$\underset{\text{Pre}}{\rightarrow}$R\textsubscript{val}} & \rotatebox{0}{S$\underset{\text{}}{\rightarrow}$R\textsubscript{val}} &
\rotatebox{0}{S$\underset{\text{Pre}}{\rightarrow}$R\textsubscript{test}} & \rotatebox{0}{S$\underset{\text{}}{\rightarrow}$R\textsubscript{test}} & \rotatebox{0}{Ar$\underset{\text{Pre}}{\rightarrow}$Cl} &\rotatebox{0}{Ar$\underset{\text{}}{\rightarrow}$Cl} & \rotatebox{0}{Pr$\underset{\text{Pre}}{\rightarrow}$Ar} & \rotatebox{0}{Pr$\underset{\text{}}{\rightarrow}$Ar}\\\toprule
$\times$ & $\times$ & $\times$ & 50.5 & 77.9 & 49.4 & 78.7 & 44.0 & 52.5 & 51.1 & 58.1\\
$\mathcal{S}$ & $\times$ & $\times$ & 52.3 & 77.5 & {51.3} & 82.5 & {46.0} & 54.1 & {54.1} & 58.8 \\
$\mathcal{S}$ & $\mathcal{S}$ & $\times$ & 52.3 & 77.9 & {51.3}  & 83.8 & {46.0} & 54.6 & {54.1} & 61.8\\
$\mathcal{S}$ & $\mathcal{T}$ & $\times$ & 52.3 & 66.3 & {51.3}  & 67.0 & {46.0} & 38.6 & {54.1} & 47.5\\
$\mathcal{S}$ & $\mathcal{S},\mathcal{T}$ & $\times$ & 52.3 & 65.7 & {51.3}  & 67.8 & {46.0} & 47.4 & {54.1} & 53.6\\
$\mathcal{S}$ & $\mathcal{S}$ & SL & 52.3 & 81.4 & {51.3}  & 85.4 & {46.0} & 56.7 & {54.1} & 64.1\\
$\mathcal{S}$ & $\mathcal{S}$ & DE & 52.3 & 85.0 & {51.3}  & 89.5 & {46.0} & 57.4 & {54.1} & 65.1\\\hline\Tstrut
$\mathcal{S}$ & $\mathcal{S}$ & DE+SL & \textbf{52.3} & \textbf{85.2} &\textbf{ 51.3} & \textbf{89.8} &\textbf{ 46.0} & \textbf{58.3} & \textbf{54.1} & \textbf{67.0}\\
\toprule
\end{tabular}}

\caption{Ablation study using ResNet-101 on the VisDA 2017 S$\rightarrow$R task (mean class accuracy) for both validation and test set as well as ResNet-50 on the Office-Home Ar$\rightarrow$Cl and Pr$\rightarrow$Ar tasks (accuracy). Transfer tasks marked with \raisebox{0.4em}{$\underset{\text{Pre}}{\rightarrow}$} indicate results after the source only pretraining and before the adaptation step. The last table row represents our full UBR$^2$S method. \label{table:ablation_full}}
\end{table}

\clearpage
\bibliography{main}
\end{document}